\pdfoutput=1
\documentclass{amia}
\usepackage{graphicx}
\usepackage[labelfont=bf]{caption}
\usepackage[superscript,nomove]{cite}
\usepackage{color}
\usepackage{url}
\usepackage{multicol}
\usepackage{multirow}
\usepackage{subfig}
\usepackage[symbol]{footmisc}
\usepackage[flushleft]{threeparttable}

\begin{document}

\title{Visual Explanations From Deep 3D Convolutional Neural Networks for Alzheimer's Disease Classification}

\author{Chengliang Yang, Anand Rangarajan, Ph.D., and Sanjay Ranka, Ph.D.}
\institutes{
	Dept. of Computer \& Information Science \& Engineering \\
	University of Florida, Gainesville, FL 32611, USA \\
	ximen14@ufl.edu, anand@cise.ufl.edu, ranka@cise.ufl.edu
}

\maketitle
\noindent{\bf Abstract}

\textit{We develop three efficient approaches for generating visual explanations from 3D convolutional neural networks (3D-CNNs) for Alzheimer's disease classification. One approach conducts sensitivity analysis on hierarchical 3D image segmentation, and the other two visualize network activations on a spatial map. Visual checks and a quantitative localization benchmark indicate that all approaches identify important brain parts for Alzheimer's disease diagnosis. Comparative analysis show that the sensitivity analysis based approach has difficulty handling loosely distributed cerebral cortex, and approaches based on visualization of activations are constrained by the resolution of the convolutional layer. The complementarity of these methods improves the understanding of 3D-CNNs in Alzheimer's disease classification from different perspectives.}
\section{Introduction}
For years, medical informatics researchers have pursued data-driven methods to automate disease diagnosis procedures for early detection of many deadly diseases. Treatment of Alzheimer's disease, which has become the sixth leading cause of death in the United States \cite{xu2016mortality}, is one of the conditions that could benefit from computer-aided diagnostic techniques. A particular challenge of Alzheimer's disease is that it is difficult to detect in early stages before mental decline begins. But medical imaging holds promise for earlier diagnosis of Alzheimer's disease \cite{mckhann2011diagnosis}. Magnetic resonance imaging (MRI), computed tomography (CT), and positron emission tomography (PET) scans contain information about the effects of Alzheimer's disease on the brain’s structure and functioning. But analyzing such scans is very time consuming for doctors and researchers because each scan contains millions of voxels. 

Deep learning systems are one potential solution for processing medical images automatically to make diagnosing Alzheimer's disease more efficient. 3D convolutional neural networks (3D-CNN), taking only MRI brain scans and disease labels as input and trained end-to-end, are reported to be on par with the performance of traditional diagnostic methods in Alzheimer's disease classification \cite{khvostikov20183d,korolev2017residual}. However, the process that 3D-CNNs use to arrive at their conclusions lacks transparency and cannot straightforwardly provide reasoning and explanations as human experts do in diagnosis. It is therefore difficult for human practitioners to trust such systems in evidence-centered areas like medical research.

The goal of this study is to break into the black box of 3D-CNNs for Alzheimer's disease classification. Particularly, we develop techniques to produce visual explanations that can indicate a 3D-CNN's spatial attention on MRI brain scans when making predictions. Our approaches give diagnosticians a better understanding of the behaviors of 3D-CNNs and provide greater confidence about integrating them into automated Alzheimer's disease diagnostic systems. In summary, the contributions of this study are as follows: 
\begin{itemize}
	\item
	We propose a hierarchical MRI image segmentation based approach for sensitivity analysis of 3D-CNNs, which can discriminate the importances of homogeneous brain regions at different levels for Alzheimer's disease classification.
	\item
	We extend two state-of-the-art approaches for explaining CNNs in 2D natural image classification to 3D MRI images, which can track the spatial attention of 3D-CNNs when predicting Alzheimer's disease. 
	\item
	We compare the developed approaches qualitatively by examining the visual explanations generated. We also conduct quantitative comparisons for their ability to localize important parts of the brain in diagnosing Alzheimer's disease.
\end{itemize}
The rest of the paper is organized as follows. Section \ref{sec2} surveys related work for this study. Section \ref{sec3} describes the methods development, data, and experimental setup. Section \ref{sec4} presents the qualitative and quantitative comparisons for proposed methods. Section \ref{sec5} presents study conclusions. 
\section{Related Work}\label{sec2}
Works that are closely connected to this study are divided into three parts: 3D-CNNs for Alzheimer's disease classification, brain MRI segmentation, and visualizing and understanding CNNs for natural image classification.
\paragraph{3D-CNNs for Alzheimer's Disease Classification}
There are two major methods for using 3D convolutional neural networks for Alzheimer's disease classification from brain MRI scans. One uses 3D-CNNs to automatically extract generic features from MRIs and build other classifiers on top of them \cite{suk2014hierarchical,hosseini2016alzheimer}. The other trains the 3D-CNNs in an end-to-end manner that only takes MRI scans and labels as input \cite{korolev2017residual,khvostikov20183d}. Both approaches achieve comparable performance \cite{khvostikov20183d}. The user has more control over the first method and thus can understand it better. The latter needs little input from humans so that it is easier to use. 
\paragraph{Brain MRI Segmentation}
As one of the fundamental problems in neuroimaging, brain segmentation is the building block for many Alzheimer's disease diagnosis methods. Semantic segmentation methods such as FreeSurfer \cite{fischl2012freesurfer} enable brain volume calculations from MRI scans of Alzheimer's disease subjects \cite{mulder2014hippocampal}. Unsupervised hierarchical segmentation methods detect homogeneous regions and separate them from coarse to finer levels, providing more flexibility for multilevel analysis than the one-level semantic segmentation \cite{corso2008efficient,yang2016supervoxel}.
\paragraph{Visualizing and Understanding CNNs for Natural Image Classification}
To explain the superior image classification performance for 2D-CNNs, researchers incorporate the spatial structure of the convolutional layer to visualize the discriminative object from activation maps \cite{zhou2016learning,selvaraju2016grad}. Sensitivity analysis by measuring the change of output class probability due to perturbed input is another popular method because it is not subject to the architectural constraints of CNNs. LIME, or local interpretable model-agnostic explanations \cite{ribeiro2016should}, is a regression-based sensitivity analysis approach that examines perturbed superpixels to make CNN results more interpretable. The perturbed superpixels could be further learned to be more semantically meaningful \cite{fong2017interpretable,yang2018global}. All these methods create a 2D spatial heatmap as a visual explanation that indicates where the CNN has focused to make its predictions. These can be extended to 3D for Alzheimer's disease classification. 

\section{Method}\label{sec3}
In this section, we describe the methods that can produce visual explanations of predictions of Alzheimer's disease from brain MRI scans by deep 3D convolutional neural networks (3D-CNNs). First, we summarize the deep learning models we deploy for the Alzheimer's disease classification task. Then, we present the brain MRI data for the study and describe how we use the data in experiments. Finally, we introduce the three approaches that we develop for explaining the 3D-CNNs, which are sensitivity analysis by 3D ultrametric contour map (SA-3DUCM), 3D class activation mapping (3D-CAM), and 3D gradient-weighted class activation mapping (3D-Grad-CAM).
\subsection{Architecture of Deep 3D Convolutional Neural Networks}
The architecture of the deep 3D convolutional neural networks (3D-CNN) for Alzheimer's disease classification in this study are based on the network architectures proposed by Korolev et al.\cite{korolev2017residual}. Particularly, two types of 3D-CNNs are built for classifying brain MRI scans from an Alzheimer's disease cohort (AD) and a normal cohort (NC). The design ideas for both types of 3D-CNNs are rooted in successful 2D natural image classification models, specifically, VGGNet, the Very Deep Convolutional Networks \cite{simonyan2014very}, and ResNet, the Deep Residual Networks \cite{he2016deep}. 

\begin{figure*}[htbp]
	\begin{center}
		\includegraphics[width=130mm]{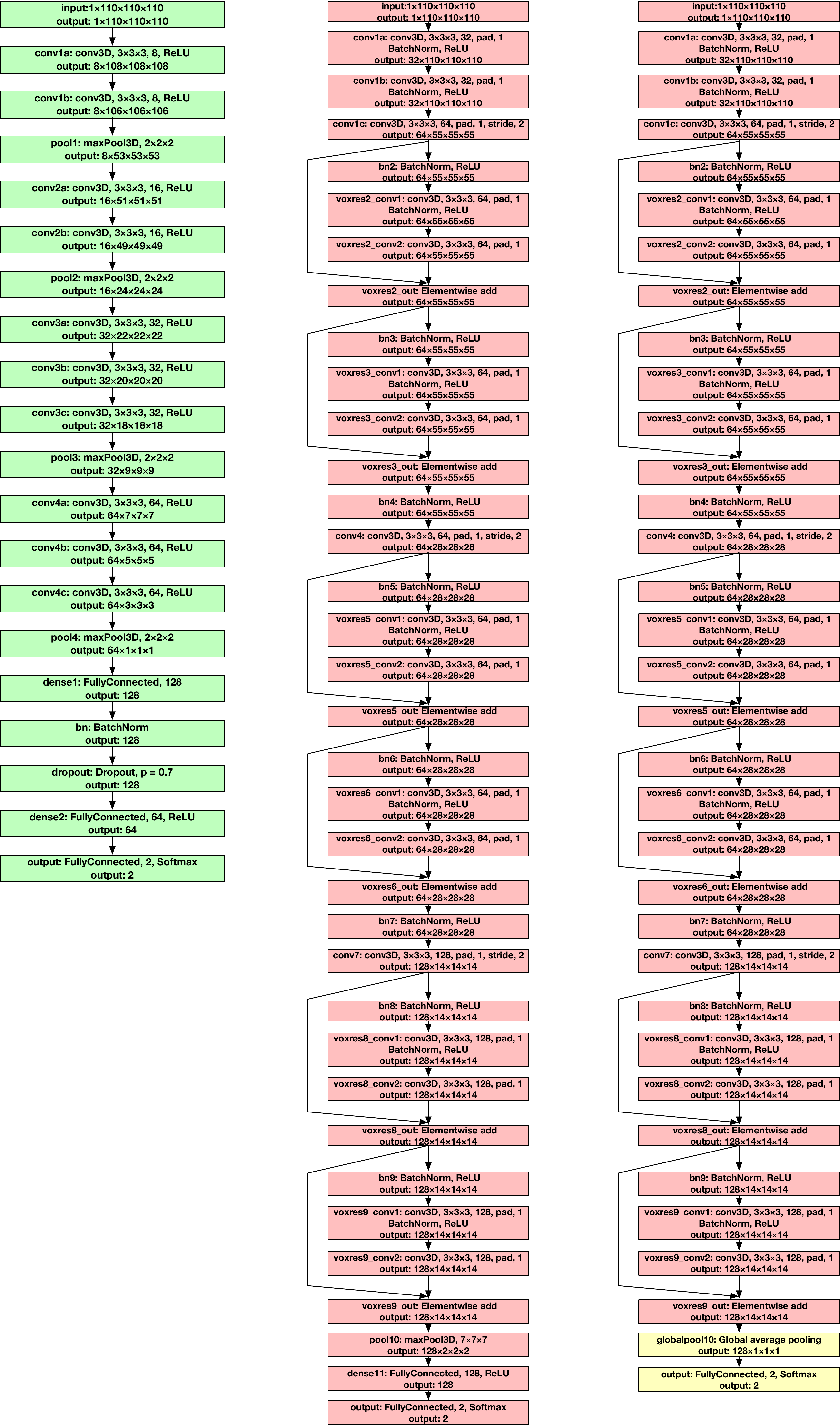}
		
		\caption{\textbf{Left:} The architecture of 3D-VGGNet; \textbf{Middle:} The architecture of 3D-ResNet; \textbf{Right:} The modified architecture of 3D-ResNet with global average pooling layer, 3D-ResNet-GAP, to produce 3D class activation mapping (3D-CAM). The only difference is that a global average pooling layer directly outputs to the softmax output layer (yellow boxes), replacing the original max pooling and fully connected layers.}\label{arch}
	\end{center}
	
\end{figure*}
\paragraph{3D Very Deep Convolutional Networks (3D-VGGNet)}
VGGNet stacks many layer blocks containing narrow convolutional layers followed by max pooling layers. The 3D very deep convolutional network (3D-VGGNet) \cite{korolev2017residual} for Alzheimer's disease classification is a direct application of this idea to 3D brain MRI scans. It contains four blocks of 3D convolutional layers and 3D max pooling layers, followed by a fully connected layer, a batch normalization layer \cite{ioffe2015batch}, a dropout layer \cite{srivastava2014dropout}, another fully connected layer, and the softmax output layer to produce the probabilities of disease in the Alzheimer's disease cohort (AD) and the normal cohort (NC). The full network architecture of 3D-VGGNet is visualized in Figure  \ref{arch} (left). To optimize model parameters, the ADAM optimizer \cite{kingma2014adam} is used with a learning rate of 0.000027, a batch size of 5, and 150 training epochs. The two-class cross-entropy calculated from the probabilities output by the softmax layer and the ground-truth labels are used as loss functions. 
\paragraph{3D Deep Residual Networks (3D-ResNet)}
Residual network is the most important building block of the state-of-the-art of 2D natural image classification \cite{he2016deep,xie2017aggregated}. 3D deep residual networks (3D-ResNet) \cite{korolev2017residual} for Alzheimer's disease classification prove their effectiveness in the 3D domain. We deploy this important type of 3D-CNN in this study and try to explain its predictions. Specifically, a six-residual-block architecture is built. Each residual block consists of two 3D convolutional layers with 3 $\times$ 3 $\times$ 3 filters that have a batch normalization layer and a rectified-linear-unit nonlinearity layer (ReLU) \cite{nair2010rectified} between them. Skip connections (identity mapping of a residual block) add a residual block element-by-element to the following residual block, explicitly enabling the following block to learn a residual mapping rather than a full mapping. This eases the learning process for deeper architectures and results in better performance. The full architecture of 3D-ResNet is depicted in Figure \ref{arch} (middle). For optimization, Nesterov accelerated stochastic gradient descent \cite{nesterov1983method} is used. Optimization parameters are set as 0.001 for learning rate, 3 for batch size, and 150 for training epochs. The same loss function as 3D-VGGNet, the two-class cross-entropy function, is used.

\subsection{Data and Experiment Setup}\label{cv}
Brain MRI scans from the Alzheimer's Disease Neuroimaging Initiative \footnote{Data used in preparation of this article were obtained from the Alzheimer's Disease Neuroimaging Initiative (ADNI) database (adni.loni.usc.edu). As such, the investigators within the ADNI contributed to the design and implementation of ADNI and/or provided data but did not participate in analysis or writing of this report. A complete listing of ADNI investigators can be found at: \url{http://adni.loni.usc.edu/wp-content/uploads/how_to_apply/ADNI_Acknowledgement_List.pdf}} (ADNI) \cite{mueller2005alzheimer} are used for this study. Specifically, we used data from the "spatially normalized, masked, and N3-corrected T1 images" category to train the 3D-VGGNet and 3D-ResNet models to classify MRI scans from the Alzheimer's disease cohort (AD) and the normal cohort (NC). Each brain MRI scan is a 3D tensor of intensity values with size 110 $\times$ 110 $\times$ 110. As one subject could have more than one MRI scan in the database, to avoid potential information leak between the training and testing dataset, we only include the earliest MRI associated with each subject for this study. As a result, 47 MRI scans from the Alzheimer's disease cohort (AD) and 56 MRI scans from the normal cohort (NC) are selected for this study. We randomly set aside eight MRI scans (5 AD, 3 NC) for later visual explanation analysis. The rest of the dataset is used for training and testing the deep 3D convolutional neural networks (3D-CNNs). 

For training and testing the 3D-VGGNet and 3D-ResNet models, we conduct five-fold cross-validation for five different splits of the dataset, totaling 25 training and testing rounds. As the batch size parameters are chosen as small numbers for both models (five for 3D-VGGNet and three for 3D-ResNet), we enforce that each batch in training contains samples from both the Alzheimer's disease cohort (AD) and normal cohort (NC) to stabilize the training process by avoiding biased loss. 

\subsection{Explaining the 3D-CNNs}
In this section, we describe the methods that we develop for explaining the predictions of the 3D-CNNs in detail. We first revisit a baseline method using sensitivity analysis that can shed light on 3D-CNNs' attention \cite{korolev2017residual}. Then we show how we used an unsupervised 3D hierarchical volumetric image segmentation approach, the 3D ultrametric contour map (3D-UCM) \cite{yang2016supervoxel}, to improve the baseline, which we call sensitivity analysis by 3D ultrametric contour map (SA-3DUCM). Next, we describe how the successful 2D visual explanation method, class activation mapping (CAM) \cite{zhou2016learning}  and its generalization, gradient-weighted class activation mapping (Grad-CAM) \cite{selvaraju2016grad}, are extended to 3D to explain predictions from 3D MRI scans. We call the two extended approaches 3D-CAM and 3D-Grad-CAM, respectively. As we mentioned, there are two major ways to explain the predictions of deep convolutional neural networks. One way applies perturbations to data and conducts sensitivity analysis. The baseline method and proposed SA-3DUCM approach belong to this category. The other way utilizes the architectural properties of CNNs to heuristically track the attention of neural networks. 3D-CAM and 3D-Grad-CAM fall into this category.
\paragraph{Baseline Approach}
A baseline approach is proposed alongside the work of 3D-VGGNet and 3D-ResNet \cite{korolev2017residual} to shed light on 3D-CNN's attention when classifying MRI scans. To be specific, for every voxel in the MRI scan, its 7 $\times$ 7 $\times$ 7 neighborhood is occluded from the image, and then the 3D-CNN re-evaluates the probability of Alzheimer's disease from the partially occuluded image. The change of probability is used as the importance of that voxel. More formally, for the brain MRI volume $V$ and each voxel of $V$ at $(x,y,z)$, we occlude the neighborhood $V_{x-3:x+3,y-3:y+3,z-3:z+3}$, resulting in a perturbed MRI volume occluded around $(x,y,z)$, denoted by $OV_{(x,y,z)}$. We want to measure the change of probability of Alzheimer's disease of $OV_{(x,y,z)}$, predicted by the 3D-CNN, compared to the original volume $V$. This change is assigned to the voxel at $(x,y,z)$. For a 3D heatmap, $C$, of the same size as $V$, to store these changes of probabilities as the importance score for all the voxels, the magnitude at $(x, y, z)$ of $C$ is calculated by 
\begin{equation}
C_{x,y,z} = |P(OV_{(x,y,z)}) - P(V)|
\end{equation}
where $P(\cdot)$ is one forward pass of the 3D-CNN to evaluate the probability of Alzheimer's disease from the MRI volumes, and $|\cdot|$ is the absolute value function. 

This approach is a direct application of the one-at-a-time sensitivity analysis at the single voxel level to test how the uncertainty of the output probability of the 3D-CNN could be assigned to different voxels of the MRI scan. This is straightforward to implement; however, this approach suffers from three important problems. First, the 7 $\times$ 7 $\times$ 7 cubical neighborhoods are not necessarily semantically meaningful and could be across different brain segments, e.g., half in cerebral cortex and half in white matter. Thus, occlusion of such an area results in an unaccountable change of output probability. Second, this approach could only capture the impact of the 7 $\times$ 7 $\times$ 7 local areas. The importances of larger or smaller areas are not tested. Third, as we evaluate a new output probability for each voxel, this approach is extremely computationally intensive. An MRI scan of size 110 $\times$ 110 $\times$ 110 has over 1 million voxels, requiring the same number of forward passes through the 3D-CNN, which could take hours even in GPU-assisted systems. 

\paragraph{Sensitivity Analysis by 3D Ultrametric Contour Map (SA-3DUCM)}
We notice that the shortcomings of the baseline approach could be overcome by using a good segmentation of the brain volume instead of the 7 $\times$ 7 $\times$ 7 local neighborhood around each voxel. Particularly, we occlude each segment in the segmentation, instead of the cubical neighborhoods, before re-evaluating the probabilities. To resolve each of the three problems of the baseline approach, the segmentation method should be semantically meaningful, hierarchical, and compact. Most specifically, to be semantically meaningful, the segmentation should separate different homogeneous parts of the brain volume well, e.g., separating cerebral cortex and white matter, so that changes of probability could be ascribed to specific segments. To be hierarchical, the segmentation method should provide a hierarchy of segmentations that capture both coarse level parts, such as the whole white matter, as well as finer level parts. In this way, we can test the importances for both small and large areas. To be compact, the segmentation method should avoid over-segmentation and generate a manageable number of segments for analysis. Thus, we can reduce the number of forward passes needed through the 3D-CNN from the number of voxels to the number of segments, which is usually three to four orders of magnitude less.

3D Ultrametric Contour Map (3DUCM) \cite{yang2016supervoxel,huang2018supervoxel} is an effective approach for unsupervised hierarchical 3D volumetric image segmentation, which is the 3D extension of the 2D state-of-the-art, Ultrametric Contour Map for natural image segmentation \cite{arbelaez2011contour}. It provides compact hierarchical segmentation of high quality. For the brain MRI volume, $V$, it could generate a hierarchy of segmentation, $H=\{H_{1},H_{2},...,H_{N}\}$, where each level $H_{n}=S_{1}^{n}\cup S_{2}^{n}\cup ...\cup S_{K_{n}}^{n}$ is a full segmentation of the volume $V$. We occlude each segment $S_{k}^{n}$ , $k=1,2,...,K_{n}$, $n=1,2,...,N$, in $V$, denoting each resulting volume by $OV_{k}^{n}$ , and re-evaluate the probability of Alzheimer's disease through one forward pass of the 3D-CNN. The change of probabilities compared to what is obtained from the original volume, $|P(OV_{k}^{n})-P(V)|$, is assigned to every voxel in $S_{k}^{n}$. Since each voxel belongs to one segment at each level of the hierarchy, each voxel gets $N$ quantities from the calculation, where $N$ is the number of levels in the segmentation hierarchy. We compute the average quantity from the $N$ quantities as the importance score for each voxel and store it in a heatmap $C$. So for a voxel of $V$ at $(x, y, z)$, assuming that it belongs to $S_{k_{n}}^{n}$, for each level of hierarchy $H_{n}$, we calculate the importance score for it as 
\begin{equation}
C_{x,y,z} = \frac{1}{N}\sum_{n=1}^{N}|P(OV_{k_{n}}^{n})-P(V)|
\end{equation}
Since the 3DUCM hierarchical segmentation usually provides homogeneous segments of the brain MRI, we expect the importance heatmap $C$ to distinguish important brain parts for Alzheimer's disease classification. In terms of computational burden, each level of the hierarchy contains at most hundreds of segments, and the hierarchy itself is no more than 20 levels. Thus, the number of forward passes needed to re-evaluate the probabilities is greatly reduced.
\paragraph{3D Class Activation Mapping (3D-CAM)}
One major problem with one-at-a-time sensitivity analysis based methods (baseline and SA-3DUCM) is that the correlations and interactions between segments of MRI volume are ignored. Although using the hierarchical segmentation method can cover most semantic segments from finer to coarser level, we cannot guarantee all combinations are tested. Therefore, we turn to methods based on the architectural properties of the 3D-CNN that directly visualize the activations of convolutional layers when predictions are made. Class activation mapping \cite{zhou2016learning} designs a global average pooling layer on top of convolutional layers in natural images classification, which enables remarkable localization performance on important objects in the images in spite of the fact that the CNN is trained on image-level labels. This fits our problem well. Our Alzheimer's disease labels (Alzheimer's disease cohort (AD) and normal cohort (NC)) are used at MRI scan level during the training of the 3D-CNNs. Our goal is to obtain visual explanations that can highlight brain parts important for Alzheimer's disease classification. Thus, extending class activation mapping to 3D provides a way to do this. 

The idea of class activation mapping is that the last convolution layer of the CNN contains the spatial information indicating discriminative regions to make classifications. To visualize these discriminative parts, class activation mapping creates a spatial heatmap out of the activations from the last convolutional layer. Specifically, class activation mapping adopts a global average pooling layer between the final convolutional layer and output layer, which enables projection of class weights of the output layer onto the activation maps in the convolutional layer. The 3D extension of class activation mapping based on 3D-ResNet is shown in Figure \ref{arch} (right). Instead of using a max pooling layer and a fully connected layer before output, the modified 3D-ResNet only uses a global average pooling layer (3D-ResNet-GAP). To be specific, for a given MRI volume $V$ and a 3D-CNN, let $f_{u}(x,y,z)$ be the activation of unit $u$ in the last convolutional layer at location $(x, y, z)$. The global average pooling for unit $u$ is $F_{u}=\frac{1}{Z}\sum_{x,y,z}f_{u}(x,y,z)$, where $Z$ is the number of voxels in the corresponding convolutional layer. As the global average pooling layer is directly connected to the softmax output layer, by the definition of the softmax function, the probability of Alzheimer's disease, $P(V)$, given by
\begin{equation}
P(V) = \frac{\exp (\sum_{u}w_{u}^{AD}F_{u})}{\exp(\sum_{u}w_{u}^{AD}F_{u})+\exp(\sum_{u}w_{u}^{NC}F_{u})}
\end{equation}
where $w_{u}^{AD}$ and $w_{u}^{NC}$ are the class weights in the output layer for the Alzheimer's disease cohort (AD) and the normal cohort (NC), respectively. We ignore the bias term here because its impact is minimal on classification performance. Essentially, $\sum_{u}w_{u}^{AD}F_{u}$ and $\sum_{u}w_{u}^{NC}F_{u}$ are the class scores for AD and NC cohorts, respectively. By extending $F_{u}$ in the class score, we have
\begin{equation}
\textrm{Score}(AD) = \sum_{u}w_{u}^{AD}F_{u} = \sum_{u}w_{u}^{AD}\frac{1}{Z}\sum_{x,y,z}f_{u}(x,y,z) = \frac{1}{Z}\sum_{x,y,z}\sum_{u}w_{u}^{AD}f_{u}(x,y,z)
\end{equation}
The $\sum_{u}w_{u}^{AD}f_{u}(x,y,z)$ part of the quantity is defined for every spatial location $(x, y, z)$ and their sum is proportional to the class score for Alzheimer's disease. As areas significantly negatively contributing to the class score are also important, we adopt the absolute value and define the class activation mapping for the AD cohort as 
\begin{equation}
\textrm{3D-CAM}_{x,y,z}(AD) = |\sum_{u}w_{u}^{AD}f_{u}(x,y,z)|
\end{equation}
which is essentially a heatmap of weighted sums of activations in every location $(x, y, z)$ and can be easily calculated by one forward pass when the volume $V$ is provided. 

Though 3D-CAM is easy to obtain, and we expect it to highlight the important spatial areas for classification, there are two potential problems with this approach. First, as we modify the 3D-CNN architecture with the global average pooling layer, we need to re-train the model, possibly affecting the classification performance. Second, the resolution of the class activation mapping is of the same size as the last convolutional layer. We need to upsample it to the original MRI scan size to identify the discriminative regions, which means we would lose some details in the resulting heatmap. One solution could be to remove more layers and build the global average pooling layers on convolutional layers with higher resolution. But this could further decrease the classification performance. 

\paragraph{3D Gradient-Weighted Class Activation Mapping (3D-Grad-CAM)}
To overcome class activation mapping's shortcoming of decreased classification performance, its generalization, gradient-weighted class activation mapping, is proposed in natural image classification \cite{selvaraju2016grad}. This approach does not need to modify the 3D-CNN's architecture and thus will do no harm to classification performance. Since no re-training is required, it is more efficient to deploy in deep learning systems. The core idea is still to identify the important activations from feature maps in convolutional layers. Using the same notation as the previous part, we first calculated the gradient of the $\textrm{Score}(AD)$ with respect to the activation of unit $u$ at location $(x, y, z)$, $f_{u}(x,y,z)$, in the last convolutional layer. Then, we use the global average pooling of the gradients, denoted by $a_{u}^{AD}$, as the importance weights for unit $u$ for the Alzheimer's disease cohort (AD). That is,
\begin{equation}
a_{u}^{AD}=\frac{1}{Z}\sum_{x,y,z}\frac{\partial \textrm{Score}(AD)}{\partial f_{u}(x,y,z)}
\end{equation}

where $Z$ is the number of voxels in the corresponding convolutional layer. Then, we combined the unit weights with the activations, $f_{u}(x,y,z)$, to get the heatmap of 3D gradient-weighted class activation mapping.
\begin{equation}
\textrm{3D-Grad-CAM}_{x,y,z}(AD) = |\sum_{u}a_{u}^{AD}f_{u}(x,y,z)|
\end{equation}
3D-Grad-CAM could be applied to a wider range of 3D-CNNs than 3D-CAM as long as the 3D-CNN has a fully convolutional layer. Also, it has been proven in 2D applications that CAM is a special case of Grad-CAM with the global average pooling layer \cite{selvaraju2016grad}. It does not require re-training so it quickly generates the 3D-Grad-CAM heatmap with just one forward pass. However, 3D-Grad-CAM still suffers from the low resolution problem because the 3D-Grad-CAM is a coarse heatmap of the same size as the last convolutional layer. We could have calculated it with gradients and activations from lower convolutional layers, but there is no guarantee that the spatial activations wouldn't change in the upper layers.

In summary, in this section, we introduce four approaches to obtain visual explanation heatmaps for predictions from 3D-CNNs. The baseline approach and sensitivity analysis by 3D ultrametric contour map (SA-3DUCM) are completely model-agnostic and can handle any type of 3D-CNNs, but they might have problems with correlations and interactions between different segments of the brain volume. 3D class activation mapping (3D-CAM) and 3D gradient-weighted class activation mapping (3D-Grad-CAM) are weighted visualizations of the activation maps in the convolutional layer, which avoids dealing with the correlations and interactions problem. However, they are limited by the low resolution of the convolutional layers. Upsampled heatmaps might not be able to provide enough detail to accurately identify important regions. For computational efficiency, the baseline approach is the slowest because it does a forward pass for every voxel. 3D-CAM only needs one forward pass to generate the heatmap, but it requires very time-consuming re-training. SA-3DUCM needs a few hundred forwarded passes. 3D-Grad-CAM is the best because it does not require re-training and only needs one forward pass when generating the heatmap. In the next section, we will compare the models' performances in identifying of discriminative brain parts for Alzheimer's disease classification from MRI scans.
\section{Results}\label{sec4}
In this section, we will present the classification performance of 3D-CNNs, visual comparisons of the heatmaps generated by the proposed visual explanation approaches, and a quantitative benchmark for the localization ability of the heatmaps in identifying important brain parts for Alzheimer's disease classification.
\subsection{Alzheimer's Disease Classification Performance}
We compare the classification performance of four different 3D-CNNs. These include 3D-VGGNet and 3D-ResNet as described. By implementing the 3D-CAM, we have a modified 3D-ResNet with global average pooling layer (GAP) as shown in Figure \ref{arch} (right), denoted as 3D-ResNet-GAP. The counterpart for 3D-VGGNet is not included because the classification performance drops too much, compared to 3D-VGGNet. Additionally, to obtain a higher resolution 3D-CAM, we remove the layers from {\tt conv4} to {\tt voxres9\_out}, resulting in a shallow version of 3D-ResNet-GAP, which we call 3D-ResNet-Shallow-GAP. All four 3D-CNNs are trained for classifying the Alzheimer's cohort (AD) in comparison to the normal cohort (NC). Classification performance is measured by the area under the ROC curve (AUC) and classification accuracy (ACC). Cross-validation as described in Section \ref{cv} is conducted. Average AUC and ACC and their standard deviations are reported. The results are presented in Table \ref{cls}. 3D-VGGNet and 3D-ResNet achieve good classification performances. However, there is a substantial drop in performance for 3DResNet-GAP and 3D-ResNet-Shallow-GAP, which means the global average pooling layer have a negative effect on classification performance.

 \begin{table*}[t]
	\begin{center}
		\begin{tabular}{p{4cm}|p{3.5cm}p{3.5cm}}
			\hline
			\textbf{Method} & \textbf{AUC} & \textbf{ACC} \\
			\hline
			3D-VGGNet & 0.863$\pm$0.056 & 0.766$\pm$0.095 \\
			3D-ResNet & 0.854$\pm$0.079 & 0.794$\pm$0.070 \\
			3D-ResNet-GAP & 0.643$\pm$0.110 & 0.614$\pm$0.100 \\
			3D-ResNet-Shallow-GAP & 0.751$\pm$0.083 & 0.585$\pm$0.122 \\
			\hline
		\end{tabular}
	\end{center}
	\caption{Classification performance of 3D-CNNs}
	\label{cls}
\end{table*}

\subsection{Qualitative Comparison for Visual Explanations}
\begin{figure*}[!ht]
	\begin{minipage}[htp]{0.49\columnwidth}%
		\begin{center}
			\subfloat[Brain MRI with highlighted cerebral cortex, lateral ventricle, and hippocampus.]{\includegraphics[width=75mm]{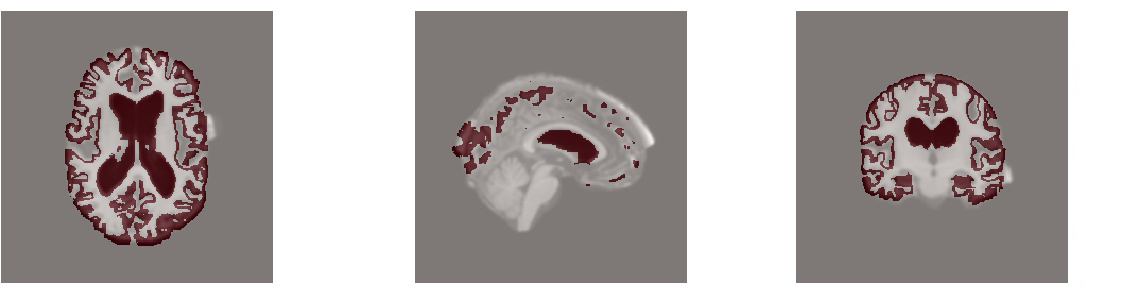}\label{gt}
				
			}
			\par\end{center}%
	\end{minipage}
	\\
	\begin{minipage}[htp]{0.49\columnwidth}%
		\begin{center}
			\subfloat[VGG-Baseline]{\includegraphics[width=75mm]{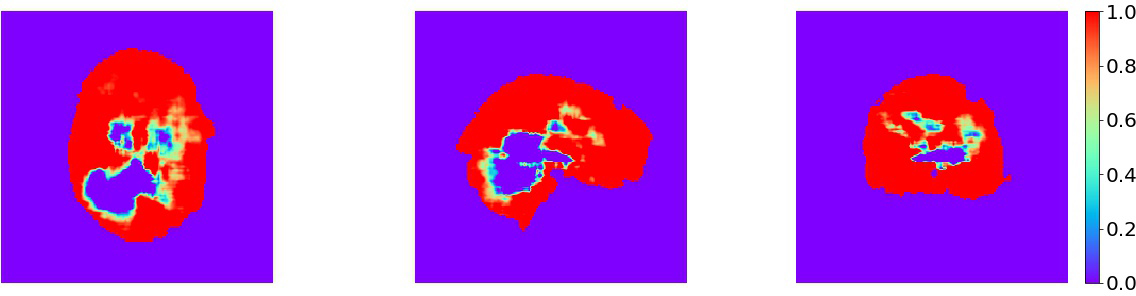}\label{vgg_baseline}
				
			}
			\par\end{center}%
	\end{minipage}
	\begin{minipage}[htp]{0.49\columnwidth}%
		\begin{center}
			\subfloat[Res-Baseline]{\includegraphics[width=75mm]{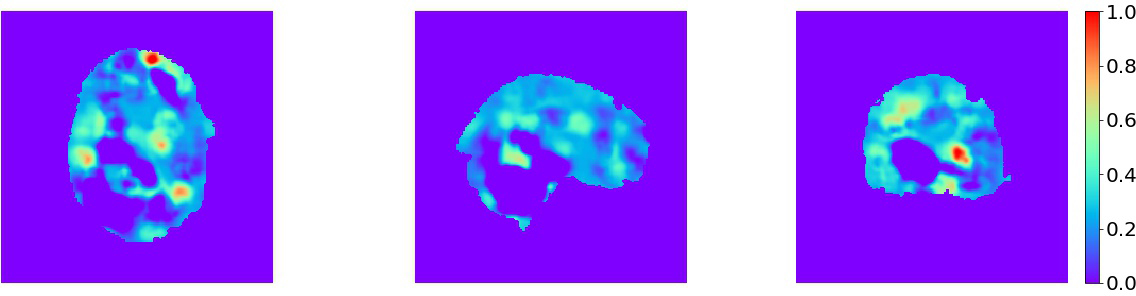}\label{resnet_baseline}
				
			}
			\par\end{center}%
	\end{minipage}
	\\
	\begin{minipage}[htp]{0.49\columnwidth}%
		\begin{center}
			\subfloat[VGG-SA-3DUCM]{\includegraphics[width=75mm]{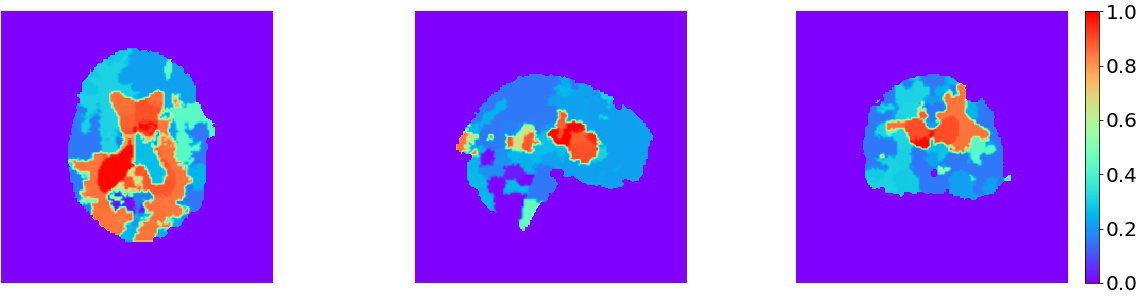}\label{vgg_3ducm}
				
			}
			\par\end{center}%
	\end{minipage}
	\begin{minipage}[htp]{0.49\columnwidth}%
		\begin{center}
			\subfloat[Res-SA-3DUCM]{\includegraphics[width=75mm]{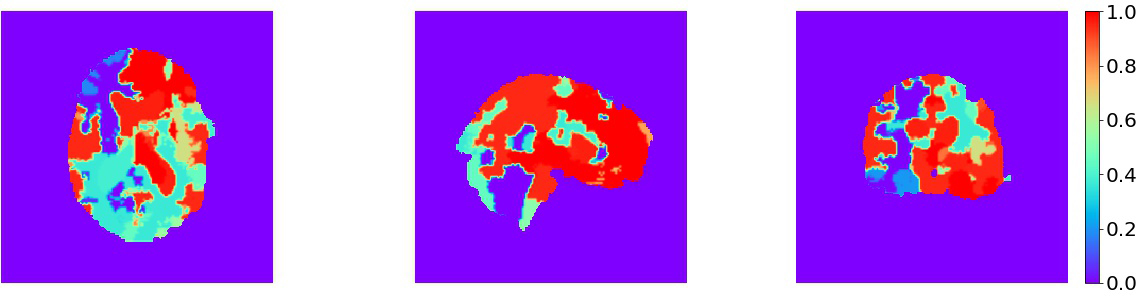}\label{resnet_3ducm}
				
			}
			\par\end{center}%
	\end{minipage}
	\\
	\begin{minipage}[htp]{0.49\columnwidth}%
		\begin{center}
			\subfloat[Res-3D-CAM]{\includegraphics[width=75mm]{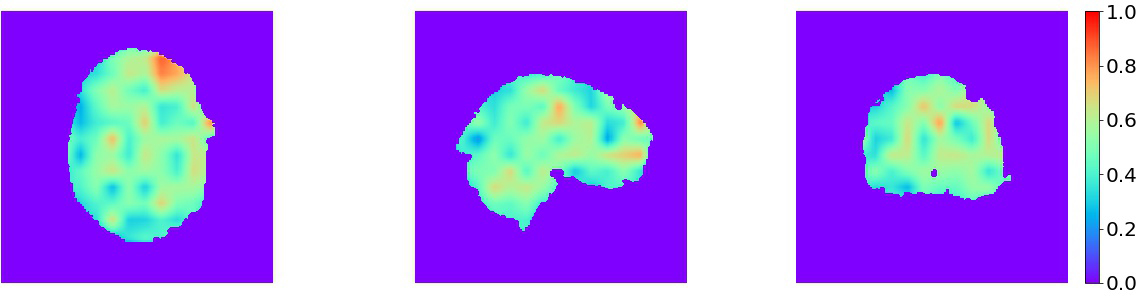}\label{resnet_cam}
				
			}
			\par\end{center}%
	\end{minipage}
	\begin{minipage}[htp]{0.49\columnwidth}%
		\begin{center}
			\subfloat[Res-3D-CAM-Shallow]{\includegraphics[width=75mm]{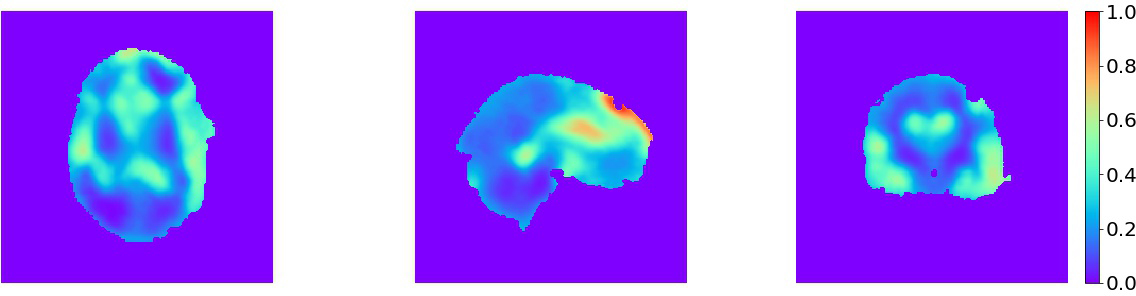}\label{resnet_shallow_cam}
				
			}
			\par\end{center}%
	\end{minipage}
	\\
	\begin{minipage}[htp]{0.49\columnwidth}%
		\begin{center}
			\subfloat[VGG-3D-Grad-CAM]{\includegraphics[width=75mm]{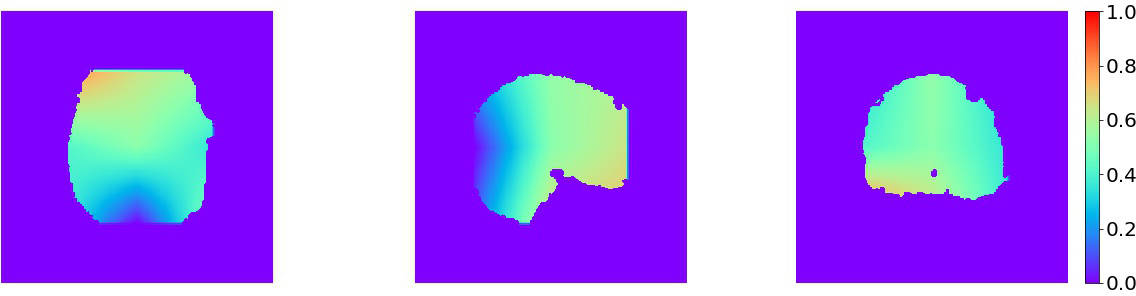}\label{vgg_grad_cam}
				
			}
			\par\end{center}%
	\end{minipage}
	\begin{minipage}[htp]{0.49\columnwidth}%
		\begin{center}
			\subfloat[Res-3D-Grad-CAM]{\includegraphics[width=75mm]{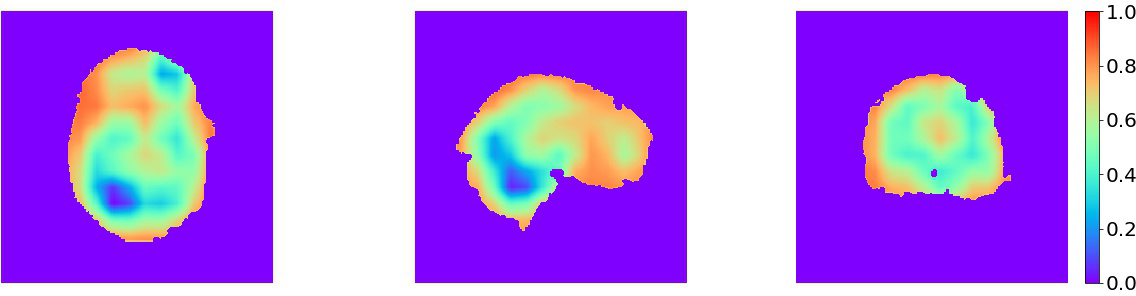}\label{resnet_grad_cam}
				
			}
			\par\end{center}%
	\end{minipage}
	\\
	\begin{minipage}[htp]{0.49\columnwidth}%
		\begin{center}
			\subfloat[VGG-3D-Grad-CAM-Shallow]{\includegraphics[width=75mm]{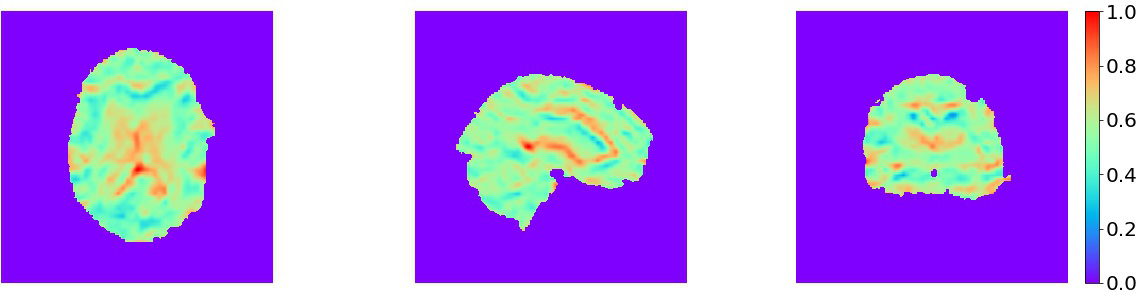}\label{vgg_grad_cam_shallow}
				
			}
			\par\end{center}%
	\end{minipage}
	\begin{minipage}[htp]{0.49\columnwidth}%
		\begin{center}
			\subfloat[Res-3D-Grad-CAM-Shallow]{\includegraphics[width=75mm]{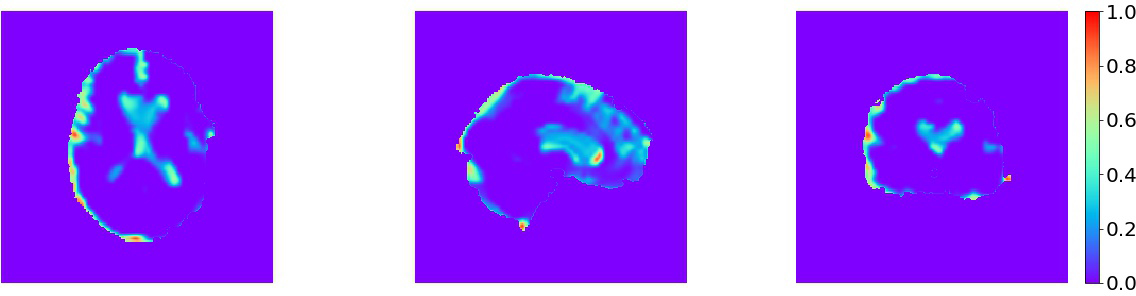}\label{resnet_grad_cam_shallow}
				
			}
			\par\end{center}%
	\end{minipage}
	\caption{Horizontal, sagittal, and coronal view of the brain MRI and the visual explanation heatmaps.}\label{comp}
\end{figure*}

To visually check the quality of heatmaps generated by the introduced visual explanation methods, we take one MRI scan from the set-aside data for visual explanation analysis and present the heatmap from the horizontal, sagittal, and coronal sections. For comparison, we present the input brain MRI volume (Figure \ref{gt}) with highlighted areas of cerebral cortex, lateral ventricle, and hippocampus. These parts are believed to be important for Alzheimer's disease diagnosis by physicians \cite{juottonen1999comparative,mu1999quantitative}. The ground-truth cerebral cortex, lateral ventricle, and hippocampus regions are segmented by the FreeSurfer software \cite{fischl2012freesurfer}. 
\paragraph{Baseline}The resulting heatmaps are labeled as VGG-Baseline and Res-Baseline and are presented in Figure \ref{vgg_baseline} and Figure \ref{resnet_baseline}, respectively. We can see from the figures that in both situations, the baseline method does not find the important areas. The heatmaps are irregularly shaped because heterogeneous regions are used for sensitivity analysis. Overall, the baseline method fails to identify discriminative regions. 

\paragraph{SA-3DUCM} After incorporating hierarchical segmentations into sensitivity analysis, we find that the results greatly improves, compared to baseline. Figure \ref{vgg_3ducm} presents the heatmap made by applying SA-3DUCM to 3D-VGGNet (VGG-SA-3DUCM), and the heatmap in Figure \ref{resnet_3ducm} is  made by applying SA-3DUCM to 3D-ResNet (Res-SA-3DUCM). In both situations, the approach differentiates the importances of different homogeneous regions. There are clear boundaries separating the regions. The lateral ventricle area stands out as the most discriminative part. However, the cerebral cortex areas are not well identified. This is because cerebral cortex is widely and loosely distributed in the brain so the cerebral cortex is usually not segmented as one area in hierarchical segmentations. SA-3DUCM tested the importance of different segments one by one. Thus, it is not able to capture the correlations between all segments that belong to the cerebral cortex. 
\paragraph{3D-CAM} We only apply 3D class activation mapping (3D-CAM) to 3D-ResNet because 3D-VGGNet loses too much classification performance after using the global average pooling layer. The class activation mapping heatmap of 3D-ResNet-GAP is labeled as Res-3D-CAM and is presented in Figure \ref{resnet_cam}. The heatmap is blurry because it is upsampled from a 14 $\times$ 14 $\times$ 14 coarse heatmap. To get a higher resolution 3D class activation mapping heatmap, Figure \ref{resnet_shallow_cam} (Res-3D-CAM-Shallow) is obtained from 3D-ResNet-Shallow-GAP with more convolutional layers removed. It is upsampled from a 55 $\times$ 55 $\times$ 55 heatmap and thus provides more detail. It identifies the lateral ventricle and most parts of the cortex as important areas, which matches the human experts' approach. 
\paragraph{3D-Grad-CAM}The 3D gradient-weighted class activation mapping (3D-Grad-CAM) also has low resolution problems, especially when it is applied to 3D-VGGNet. Because the last convolutional layer of 3D-VGGNet is only of size 3 $\times$ 3 $\times$ 3, the resulting heatmap VGG-3D-Grad-CAM barely provides any information (Figure \ref{vgg_grad_cam}). When we apply the same approach to a lower convolutional layer, {\tt conv2b}, in 3D-VGGNet, the resulting heatmap, VGG-3D-Grad-CAM-Shallow (Figure \ref{vgg_grad_cam_shallow}), is able to highlight part of the lateral ventricle. 3D-ResNet has the same situation. Res-3D-Grad-CAM (Figure \ref{resnet_grad_cam}) and Res-3D-Grad-CAM-Shallow (Figure \ref{resnet_grad_cam_shallow}) are generated by the 3D-Grad-CAM approach applied to {\tt voxres9\_out}  (last convolutional layer) and {\tt bn4} (an intermediate convolutional layer) of 3D-ResNet. They are of size 14 $\times$ 14 $\times$ 14 and 55 $\times$ 55 $\times$ 55, respectively. Though both of them identify most of the lateral ventricle and the cerebral cortex as discriminative, Res-3D-Grad-CAM-Shallow is of higher resolution and more accurate. However, as we stated, upper convolutional layers could change the activation maps from the lower convolutional layers. Thus sometimes, we may not trust the heatmap from lower layers as a good representation of spatial attention of the 3D-CNN. 

To summarize the qualitative comparisons, SA-3UCM has the same resolution as the original MRI volume and differentiates homogeneous regions well. However, it fails to identify the correlations from the fragmented cerebral cortex segments because of the one-at-a-time process in sensitivity analysis. 3D-Grad-CAM and 3D-CAM both produce more blurry heatmaps than SA-3DUCM because of upsampling. But they are able to highlight the cerebral cortex that is loosely distributed in the brain. 
\subsection{Quantitative Comparison for Localization}
\begin{figure*}[tbp]
	\begin{center}
		\includegraphics[width=140mm]{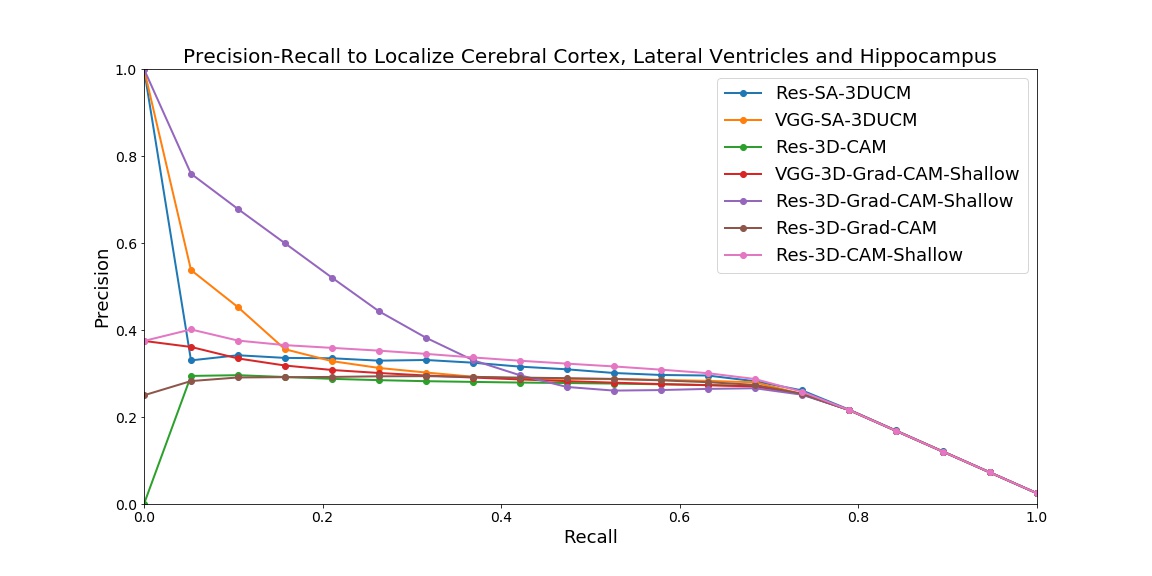}
		
		\caption{Precision-recall curve to localize cerebral cortex, lateral ventricle, and hippocampus regions using heatmaps.}\label{qf}
	\end{center}
	
\end{figure*}

Visual comparisons of the heatmap give us a general idea how well different visual explanation methods work. But we wonder how well these heatmaps could localize important regions such as cerebral cortex, lateral ventricle, and hippocampus. To quantitatively compare localization ability, we plot the precision-recall curve for the heatmaps that we have visualized in the previous section to identify cerebral cortex, lateral ventricle, and hippocampus regions from the 8 MRI scans that are set aside for visual explanation analysis. VGG-Baseline, Res-Baseline, and VGG-3D-Grad-CAM are not included because they do not generate usable heatmaps in the visual comparisons. The results are presented in Figure \ref{qf}.

From the results, we can see VGG-SA-3DUCM, Res-SA-3DUCM, and Res-3D-Grad-CAM-Shallow have high precision on the low recall end. This matches our visual comparisons as SA-3DUCM method puts the homogeneous lateral ventricle regions on top, and Res-3D-Grad-CAM-Shallow identifies cerebral cortex and lateral ventricle parts with high accuracy. However, the precision drops for all methods on the high recall end, implying no method is close to perfectly identifying all important regions. The reasons would be different. SA-3DUCM could not discriminate the cerebral cortex because of fragmented segments. 3D-CAM and 3D-Grad-CAM are limited by low resolution of the heatmaps.

Overall, both qualitative and quantitative comparisons indicate that all visual explanation methods have some limitations. The correct method may be chosen based on the specific goals. When the goal is to get the importance for a homogeneous region, SA-3DUCM is more suitable. If tracking the attention of the 3D-CNN is the goal, 3D-Grad-CAM is the preferred choice. Generally 3D-Grad-CAM is better than 3D-CAM because it does not modify the 3D-CNN architecture, requires less computation, and better localizes important regions.

\section{Conclusion and Discussion}\label{sec5}
In this study, we develop three approaches for producing visual explanations from 3D-CNNs for Alzheimer's disease classification. All approaches can highlight important brain parts for diagnosis. However, they have limitations in different aspects. The one-at-a-time sensitivity analysis procedure of SA-3DUCM is not able to handle correlated or interacting images segments, causing underestimation of attention in the loosely distributed area such as cerebral cortex in our case. 3D-CAM and 3D-Grad-CAM build heatmaps from convolutional layer activations that have lower resolution than the original MRI scan, resulting in loss of details and decreased localization accuracy. Therefore, we suggest users choose the right approach based on their use cases for MRI analysis.

Though all approaches are developed for Alzheimer's disease classification, they are generic enough for other type of 3D image analysis. SA-3DUCM is completely model agnostic and can adapt to any classifiers taking 3D volumetric images as input. 3D-CAM and 3D-Grad-CAM can work on any deep learning model that has a 3D convolutional layer. They could be applied to other types of 3D medical images or even video analysis.

One common limitation of these approaches is that the visual explanation is still one step away from fully understanding the 3D-CNN. Human experts measure cerebral cortex thickness as a biomarker for diagnosis \cite{fischl2000measuring}. In the generated visual explanations, there is no such explicit summarized representation on top of the visual attention from the cerebral cortex. This leads to our future work of explicit biomarker representation learning from medical imaging to fully interpret the 3D-CNNs.

\section*{\uppercase{Acknowledgments}}
This work is partially supported by NSF 1743050 to A.R. and S.R..

\makeatletter
\renewcommand{\@biblabel}[1]{\hfill #1.}
\makeatother

\bibliographystyle{unsrt}
\bibliography{amia}

\end{document}